\documentclass[10pt]{article} 
\usepackage[preprint]{tmlrheiko}

\usepackage{amsmath,amsfonts,bm}

\newcommand{\fwdparam}{\phi}

\newcommand{\bwdparam}{\theta}

\newcommand{\sharedparam}{\eta}
\newcommand{\allparam}{\lambda}
\newcommand{\zparam}{\psi}

\def\eqref#1{equation~\ref{#1}}

\def\1{\bm{1}}

\DeclareMathAlphabet{\mathsfit}{\encodingdefault}{\sfdefault}{m}{sl}
\SetMathAlphabet{\mathsfit}{bold}{\encodingdefault}{\sfdefault}{bx}{n}

\DeclareMathOperator*{\argmin}{arg\,min}

\usepackage{hyperref}
\usepackage{url}

\usepackage[utf8]{inputenc} 
\usepackage[T1]{fontenc}    
\usepackage{booktabs}       
\usepackage{amsfonts}       
\usepackage{nicefrac}       
\usepackage{microtype}      
\usepackage{xcolor, soul}
\usepackage{multirow}

\usepackage{amsmath, amssymb, cancel}       
\usepackage{cancel}
\usepackage{tabularx}
\usepackage{graphicx}
\usepackage{algorithm}
\usepackage{algpseudocode}

\title{A Variational Perspective on Generative Flow Networks}

\author{\name Heiko Zimmermann \email h.zimmermann@uva.nl\\
    \addr Amsterdam Machine Learning Lab,\\
    University of Amsterdam
    \AND
    \name Fredrik Lindsten \email fredrik.lindsten@liu.se\\
    \addr Division of Statistics and Machine Learning,\\
    Link\"{o}ping University
    \AND
    \name Jan-Willem van de Meent \email j.w.vandemeent@uva.nl\\
    \addr Amsterdam Machine Learning Lab,\\
    University of Amsterdam
    \AND
    \name Christian A. Naesseth \email c.a.naesseth@uva.nl\\
    \addr Amsterdam Machine Learning Lab,\\
    University of Amsterdam\\
}

\begin{document}

\maketitle

\begin{abstract}
Generative flow networks (GFNs) are a class of models for sequential sampling of composite objects, which approximate a target distribution that is defined in terms of an energy function or a reward.
GFNs are typically trained using a flow matching or trajectory balance objective, which matches forward and backward transition models over trajectories. In this work, we define variational objectives for GFNs in terms of the Kullback-Leibler (KL) divergences between the forward and backward distribution. 
We show that variational inference in GFNs is equivalent to minimizing the trajectory balance objective when sampling trajectories from the forward model. We generalize this approach by optimizing a convex combination of the reverse- and forward KL divergence. 
This insight suggests variational inference methods can serve as a means to define a more general family of objectives for training generative flow networks, for example by incorporating control variates, which are commonly used in variational inference, to reduce the variance of the gradients of the trajectory balance objective. We evaluate our findings and the performance of the proposed variational objective numerically by comparing it to the trajectory balance objective on two synthetic tasks.
\end{abstract}

\section{Introduction}
Generative flow networks (GFNs) \citep{bengio_flow_2021, bengio_gflownet_2021} have recently been proposed as a computationally efficient method for sampling composite objects such as molecule strings \citep{bengio_flow_2021}, DNA sequences \citep{jain_biological_2022} or graphs \citep{deleu_bayesian_2022}. To generate such objects, GFNs sample a trajectory along a directed acyclic graph (DAG) in which edges correspond to actions that modify the object. A trajectory sequentially constructs an object by transitioning from a root node (initial object, or null state) to a terminating node (final composite object) which is scored according to a reward signal. 
While sampling sequences of actions has been well studied in the reinforcement learning literature \citep{sutton_reinforcement_2018}, the objective is typically to find a policy which maximizes the expected reward of the trajectory.
By contrast, GFNs are trained to learn a policy that solves a planning-as-inference problem \citep{toussaint2006probabilistic} by learning a distribution over trajectories ending in a terminating state with probability proportional to the reward assigned to it. This is done by optimizing objectives which aim to satisfy a flow matching or detailed balance condition \citep{bengio_flow_2021}. \citet{malkin_gflownets_2022} has since found that these objectives are prone to ineffective credit propagation across trajectories and proposes an alternative objective based on a trajectory balance (TB) condition to alleviate these problems. Most recently,  \citet{madan_learning_2022} proposed an objective that can be optimized on partial trajectories and \citep{do_improving_2022} proposed an optimal-transport-based objective to further improve generalization and exploration.

A positive reward function can be interpreted as an unnormalized distribution, which one wishes to generate samples from. In this view, we are interested in sequentially sampling form a factorized joint distibution, such that the marginal distribution of the final state is approximately equal to the corresponding normalized distribution.
Generating approximate samples from an unnormalized target distribution is a common task in probabilistic inference, for which many methods have been developed. Examples include methods based on MCMC \citep{homan_no-u-turn_2014, salimans_markov_2015, li_approximate_2017, hoffman_learning_2017, naesseth_markovian_2021, zhang2022transport}, importance sampling  \cite{neal_annealed_2001, del_moral_sequential_2006,naesseth2019esmc} and variational inference \citep{blei2017variational,naessethLRB2018,maddison2017filtering,anh2018autoencoding, zimmermann_nested_2021}.
Recent work on GFNs by \citet{zhang_generative_2022} takes a similar view by treating the reward function as an energy-based model, which can be trained to maximize the data likelihood following a contrastive divergence-based approach \citep{hinton_training_2002}, while the forward- and backward transition models are trained by optimizing the TB objective.

In this work we show that, in certain settings, optimizing the TB objective is indeed equivalent to optimizing a forward- or reverse Kullback-Leibler divergence. To this end we compare the TB objective when optimized with samples generated form the forward transition model, backward transition model, or a mixture of both, to training with a corresponding variational objective, which takes the form of a convex combination of a forward- and reverse Kullback-Leibler divergence.
We identify cases in which the TB objective is equivalent to the corresponding variational objective and leverage this insight to employ variance reduction techniques from variations inference. Finally, we run experiments, to evaluate our theoretical findings and the empirical performance of the trajectory balance and corresponding variational objective.

\paragraph{Related Work}
Recent work by \citet{zhang_unifying_2022} identifies equivalences between GFNs and certain classes of generative models. The authors observe that hierarchical variational auto-encoders are equivalent to a special class of GFNs, and that training hierarchical latent variable models with the forward KL divergence between the full backward- and forward transition model of the GFN is equivalent to training a hierarchical VAE by maximizing its ELBO. 

In concurrent and independent work, \citet{malkin_gflownets_2022} derive the same equivalences between optimizing the TB objective and forward- and reverse KL divergence that we establish in this work. The difference with our work is that we propose a novel composite objective based on a convex combination of the reverse and forward Kullback-Leibler divergences. Furthermore, we discuss and study this objective in context of learning energy-based models. Finally, we also study the differences between variational inference and trajectory balance optimization when the forward and backward trajectory distributions share parameters.


\section{Generative Flow Networks}
\label{sec:background}
Generative flow networks \citep{bengio_flow_2021, bengio_gflownet_2021} 
generate trajectories $\tau = (s_0, s_1, \ldots, s_T, s_f)$ along the edges of a directed acyclic graph $G = (\mathcal{S},E)$. Each trajectory starts in the root, $s_0$, and terminates in a \emph{terminating state}, $s_T$, before transitioning to a special \emph{final state}, $s_f$, which is the single leaf node of $G$.
A non-negative reward signal $R(s_T)$ is assigned to each terminating state $s_T$. The task is to learn a sampling procedure, or flow, for simulating trajectories, such that the marginal distribution of reaching the terminating state $s_T$ is proportional to $R(s_T)$.
We adopt the convention that $s_f = s_{T+1}$. The structure of the DAG imposes a partial order, $<$, on states $s, s' \in \mathcal{S}$ such that $s < s'$ if $s$ is an ancestor of $s'$. Hence, any trajectory satisfies $s_j$ < $s_k$ for $0 \leq j < k \leq T+1$ and consequently does not contain loops.

\subsection{Trajectory Flows}
A \emph{trajectory flow} is a non-negative function $F_G: \mathcal{T} \to \mathbb{R}^{+}$ on complete trajectories $\mathcal{T}$, i.e. trajectories starting in a \emph{initial state} $s_0$ and ending in the final state $s_f$ associated with a DAG $G$. Below, we drop the graph subscript for notational convenience. 
A trajectory flow defines a probability measure $P$ over complete trajectories, such that for any event $A \subseteq \mathcal{T}$
\begin{align*}
    P(A) = \frac{F(A)}{Z}
    ,&&
    F(A) = \sum_{\tau \in A} F(\tau)
    ,&&
    Z = \sum_{\tau \in \mathcal{T}} F(\tau)
    ,
\end{align*}
where $Z$ can be interpreted as the total amount of flow.
The flow $F(s)$ through a state and the flow $F(s \to s')$ along an edge $(s, s')$ are denoted by
\begin{align*}
    F(s) := F(\{\tau \in \mathcal{T}: s \in \tau\})
    ,&&
    F(s \to s') := F(\{\tau \in \mathcal{T} \mid \exists t \in \mathbb{N} : s=s_t, s'=s_{t+1} \in \tau\})
    .
\end{align*}
The probability of a trajectory containing the state $s$, and the \emph{forward}- and \emph{backward transition probabilities} are denoted by
\begin{align*}
    P(s) := \frac{F(s)}{Z}
    ,&&
    P_F(s' \mid s) 
    := 
    P(s \to s' \mid s) 
    := 
    \frac{F(s \to s')}{F(s)}
    ,&&
    P_B(s \mid s') 
    := 
    P(s \to s' \mid s) 
    = 
    \frac{F(s \to s')}{F(s')}.
\end{align*}

A flow is referred to as a \emph{Markovian flow} if its corresponding probability measure satisfies $P(s \to s' \mid \tau) = P(s \to s' \mid s)$ for any consecutive states $s, s'$ and partial trajectory $\tau=(s_0, \ldots, s)$ ending in $s$. For a Markovian flow and complete trajectory $\tau \in \mathcal{T}$ we have,
\begin{align*}
    P(\tau) 
    &= 
    \prod_{t=0}^T P_F(s_{t+1} \mid s_t)
    =
    \prod_{t=0}^T P_B(s_t \mid s_{t+1})
    .
\end{align*}
For a comprehensive study of flows and generative flow networks we refer to \citet{bengio_gflownet_2021}.

\subsection{Training Generative Flow Networks} 
\label{ssec:gfn_training}
We are considering GFNs, which parameterize a Markovian flow on a DAG by modeling forward transition probabilities $P_F(s' | s; \fwdparam)$, together with a normalizing constant $Z_\zparam$ which can be interpreted as an approximation to the total amount of flow. The trajectory flow is
\begin{align*}
    F(\tau; \fwdparam, \zparam) 
    &= 
    Z_\zparam  \prod_{t=0}^{T} P_F(s_{t+1} \mid s_t; \fwdparam)
    =
    \frac{
        \prod_{t=0}^{T} F(s_t \to s_{t+1}; \fwdparam, \zparam)
    }{
        \prod_{t=0}^{T-1} F(s_{t+1}; \fwdparam, \zparam)
    }
    =
    Z_\zparam \prod_{t=0}^{T} P_B(s_{t} \mid s_{t+1}; \fwdparam),
    .
\end{align*}
For a reward function $R$, the goal is to find transition probabilities such that $P_B( s_T \mid s_{f}; \fwdparam) = R(s_T)/Z$.
%
In some scenarios we want to fix the backward transition model, e.g. a uniform distribution model can be advantageous for exploration, or parameterize it with a distinct set of parameters $\bwdparam$. 
In this case, the forward and backward transition probabilities do not correspond to the same flow and, under slight overload of notation, we refer to $P_B(s \mid s'; \bwdparam)$ as the backward transition probabilities.

\citet{bengio_flow_2021} originally proposed objectives to train GFNs based on the flow matching conditions and a detailed balance condition. \citet{malkin_trajectory_2022} observe that 
optimizing these 
may lead to inefficient credit propagation to early transitions, especially for long trajectories. To alleviate this, \citet{malkin_trajectory_2022} propose an alternative TB objective for complete trajectories 
\begin{align}
\label{eq:tb_objective}
    \mathcal{L}_{\mathrm{TB}}(\tau, \allparam) 
    = 
    \left(
    \log \frac{
        Z_\zparam \prod_{t=0}^T P_F(s_{t+1} | s_t; \fwdparam)
    }{
        R(s_T) \prod_{t=0}^{T-1} P_B(s_t | s_{t+1}; \bwdparam)
    }
    \right)^2
    = 
    \left(
    \log \frac{
        Z_\psi Q(\tau; \fwdparam)
    }{
        Z P(\tau; \bwdparam)
    }
    \right)^2,
\end{align}
where $\allparam=(\fwdparam, \bwdparam, \zparam)$ and we define
\begin{align*}
    P(\tau; \bwdparam) 
    :=
    \frac{R(s_T)}{Z}
    \prod_{t=0}^{T-1} P_B(s_{t} \mid s_{t+1}; \bwdparam), 
    &&
    Q(\tau; \fwdparam) 
    :=
    \prod_{t=0}^{T} P_F(s_{t+1} \mid s_{t}; \fwdparam).  
\end{align*}
Trajectories $\tau$ are
sampled from a proposal distribution $q$ with full support over the space of trajectories $\mathcal{T}$.
The TB objective is optimized using stochastic gradient descent. The gradient w.r.t.~all parameters $\allparam=(\fwdparam, \bwdparam, \zparam)$ is computed as the average over a batch of $S$ i.i.d.~samples. Solutions correspond to fixed points of the (negative) expected gradient
\begin{align*}
    \mathbb{E}_{\tau \sim q}
    \left[
        \frac{d}{d\allparam} \mathcal{L}_{ \mathrm{TB}}(\tau, \allparam)
    \right] 
    = 0
    .
\end{align*}
We can compute an unbiased estimate of this gradient using samples from the proposal distribution, 
\begin{align*}
    g_{\mathrm{TB}}(\allparam)
    :=
    \frac{1}{S}
    \sum_{s=1}^S
        \frac{d}{d\allparam} \mathcal{L}_{\mathrm{TB}}(\tau_s, \allparam)
    ,&&
    \tau_s \sim q
    .
\end{align*}
In section \ref{sec:method}, we show how optimizing GFNs using the TB objective corresponds to variational inference on complete trajectories. Going forward, we refer to the probability mass functions $Q(\tau; \fwdparam)$ and $P(\tau; \bwdparam)$ over complete trajectories as forward and backward model, respectively.

\section{Variational Inference}
\label{subsec:background_variational_inference}
The problem of finding corresponding forward and backward transition probabilities can alternatively be phrased as a variational inference problem. The goal is to find parameters $\fwdparam$ and $\bwdparam$ such that the difference between the forward and backward transition probabilities, measured by a suitable divergence, is minimized. Two commonly used divergence measures are the forward Kullback-Leibler divergence (FKL) and reverse Kullback-Leibler divergence (RKL),
\begin{align}
    \label{eq:vi_formulation}
    \mathcal{L}_\mathrm{RKL}(\fwdparam, \bwdparam)
    &:= 
    \mathrm{KL}(Q(\cdot\ ; \fwdparam) \mid P(\cdot\ ; \bwdparam) ) 
    = 
    \mathbb{E}_{\tau \sim Q}\left[
        \log \frac{
            Q(\tau; \fwdparam)
        }{
            P(\tau; \bwdparam)
        }
    \right]
    = 
    \mathbb{E}_{\tau \sim Q}\left[
        -\log w
    \right]
    ,
    \\
    \mathcal{L}_\mathrm{FKL}(\fwdparam, \bwdparam) 
    &:= 
    \mathrm{KL}(P(\cdot\ ; \bwdparam) \mid Q(\cdot\ ; \fwdparam)) 
    = 
    \mathbb{E}_{\tau \sim P}\left[
        \log \frac{
            P(\tau; \bwdparam)
        }{
            Q(\tau; \fwdparam)
        }
    \right]
    = 
    \mathbb{E}_{\tau \sim P}\left[
        \log w
    \right]
    ,
\end{align}
with the importance weights $w := P(\tau; \bwdparam) / Q(\tau; \fwdparam)$.
The divergences can be optimized using stochastic gradient descent with gradients estimated from samples from the forward model $Q$ and backward model $P$, respectively. In most setting, samples from $P$ are not readily available and one has to resort other techniques to generate approximate samples, e.g. using importance sampling or MCMC.

Computing the derivative of $\mathcal{L}_\mathrm{RKL}$ w.r.t.~parameters $\bwdparam$ of the backward transition model is straightforward, the dependence only appears in the log-weights. We can approximate the resulting expected gradient using $S$ samples from the forward model,
\begin{align*}
    \frac{d}{d \bwdparam}
    \mathcal{L}_\mathrm{RKL}(\fwdparam, \bwdparam)
    = 
    \mathbb{E}_{\tau \sim Q}\left[
        -\frac{d}{d \bwdparam}
        \log P(\tau; \bwdparam)
    \right]
    \approx
    g^\bwdparam_\mathrm{RKL}(\fwdparam, \bwdparam)
    :=
    \frac{1}{S}\sum_{s=1}^S
        -\frac{d}{d \bwdparam}
        \log P(\tau_s; \bwdparam)
    ,&&
    \tau_s \sim Q(\cdot; \fwdparam)
    .
\end{align*}
Similarly, the derivative of $\mathcal{L}_\mathrm{FKL}$ w.r.t.~parameters $\fwdparam$ of the forward transition model and corresponding gradient estimator $g^\fwdparam_\mathrm{RKL}$ are
\begin{align*}
    \frac{d}{d \fwdparam}
    \mathcal{L}_\mathrm{FKL}(\fwdparam, \bwdparam)
    = 
    \mathbb{E}_{\tau \sim P}\left[
        -\frac{d}{d \fwdparam}
        \log Q(\tau; \fwdparam)
    \right]
    \approx
    g^\fwdparam_\mathrm{FKL}(\fwdparam, \bwdparam)
    :=
    \frac{1}{S}\sum_{s=1}^S
        -\frac{d}{d \fwdparam}
        \log Q(\tau_s; \fwdparam)
    ,&&
    \tau_s \sim P(\cdot; \bwdparam)
    .
\end{align*}
Computing derivative of $\mathcal{L}_\mathrm{RKL}$ w.r.t.~$\fwdparam$ and derivative of $\mathcal{L}_\mathrm{FKL}$ w.r.t.~$\bwdparam$ on the other hand involves computing a so-called score-function gradient,
\begin{align*}
    \frac{d}{d \fwdparam}
    \mathcal{L}_\mathrm{RKL}(\fwdparam, \bwdparam)
    &= 
    \sum_{\tau \in \mathcal{T}} 
        \log \frac{
            Q(\tau; \fwdparam)
        }{
            P(\tau; \bwdparam)
        }
        \frac{d}{d \fwdparam}
        Q(\tau ; \fwdparam)
        +
        \left(
            \frac{d}{d \fwdparam}
            \log \frac{
                Q(\tau; \fwdparam)
            }{
                P(\tau; \bwdparam)
            }
        \right)
        Q(\tau ; \fwdparam)
    \\
    &=
    \sum_{\tau \in \mathcal{T}} 
        \log \frac{
            Q(\tau; \fwdparam)
        }{
            P(\tau; \bwdparam)
        }
        Q(\tau ; \fwdparam)
        \frac{d}{d \fwdparam}
        \log Q(\tau ; \fwdparam)
        +
        Q(\tau ; \fwdparam)
        \frac{d}{d \fwdparam}
        \log Q(\tau; \fwdparam)
    \\
    &=
    \mathbb{E}_{\tau \sim Q}\left[
        \left(
            - \log w 
            + 
            1 
        \right)
        \frac{d}{d \fwdparam}
        \log Q(\tau ; \fwdparam)
    \right]
    =
    \mathbb{E}_{\tau \sim Q}\left[
            - \log w 
        \frac{d}{d \fwdparam}
        \log Q(\tau ; \fwdparam)
    \right]
\end{align*}
Importantly, we can cancel-out the additional score-function term (last equality of above equation) as $\mathbb{E}_{\tau \sim Q}[a \frac{d}{d\fwdparam} \log Q(\tau; \fwdparam)] = 0$ for any constant $a$. 
The corresponding score-function gradient estimator is thus
\begin{align*}
    g^\fwdparam_\mathrm{RKL}(\fwdparam, \bwdparam)
    :=
    \frac{1}{S}
    \sum_{s=1}^S 
        -\log w_s
        \frac{d}{d \fwdparam}
        \log Q(\tau_s ; \fwdparam)
    ,&&
    w_s 
    := 
    \frac{
            P(\tau_s; \bwdparam)
        }{
            Q(\tau_s; \fwdparam)
        }
    ,&&
    \tau_s \sim Q(\cdot; \fwdparam)
    .
\end{align*}
Analogously, we can compute a score function gradient of $\mathcal{L}_\mathrm{FKL}$ w.r.t.~$\bwdparam$ and corresponding estimator
\begin{align*}
    \mathbb{E}_{\tau \sim P}\left[
        \log w
        \frac{d}{d \bwdparam}
        \log P(\tau ; \bwdparam)
    \right]
    \approx
    g^\bwdparam_\mathrm{FKL}(\fwdparam, \bwdparam)
    :=
    \frac{1}{S}
    \sum_{s=1}^S 
        \log w_s
        \frac{d}{d \bwdparam}
        \log P(\tau_s ; \bwdparam)
    ,\qquad 
    \tau_s \sim P(\cdot; \bwdparam)
    .
\end{align*}
Score-function gradient estimators can exhibit high variance \citep{ranganath_black_2013}, which can be problematic for learning variational approximations via stochastic gradient descent, and hence it is often essential to employ additional variance reduction techniques. 

\subsection{Variance reduction techniques for score-function estimators}
\label{subsec:background_variance_reduction}
A commonly used technique to reduce the variance of score-function estimators is to use a control variate $h$ \citep{ross_simulation_1997}
to replace the gradient estimator $g$ with the modified estimator
$g^\prime = g + c(h - \mathbb{E}\bigl[h\bigr])$,
where $c$ is a scaling parameter.
Control variates leave the expected value of the gradient estimator $g$ unchanged, 
\( \mathbb{E}[g] 
    = 
    \mathbb{E}[g'],
\)
but has the potential to reduce the variance. Indeed, for a given control variate $h$ we can minimize the variance of $g^\prime$
\begin{align}
    \label{eq:estimator_with_cv}
    \mathrm{Var}[g^\prime]
    =
    \mathrm{Var}[g]
    + c^2 \mathrm{Var}[h]
    - 2c \mathrm{Cov}[g, h]
\end{align}
with respect to the scaling $c$:
\begin{align*}
    c^* = \argmin_{c} \mathrm{Var}[g']
    =
    \frac{
        \mathrm{Cov}[g^\prime, h]
    }{
        \mathrm{Var}[h]
    }.
\end{align*}

The score function $\frac{d}{d \fwdparam} \log Q(\tau; \fwdparam)$ \citep{ranganath_black_2013} is a useful and easy to compute control variate when optimizing the reverse KL divergence, which we will use as our running example. Using the score function as a control variate simplifies the expression of the resulting gradient estimator such that the scaling $c$ can simply be added to the (negative) log-importance weight, 
\begin{align*}
    \begin{split}
    g'
    &=
    \underbrace{
    \frac{1}{S}
    \sum_{i=s}^S
    -\log w_s\
    \frac{d}{d\fwdparam}
    \log Q(\tau_s; \fwdparam) 
    }_{g}
    +
    c\bigl(
    \frac{d}{d\fwdparam}
        \log Q(\tau_s; \fwdparam) 
        - \underbrace{\mathbb{E}
        \left[
            \frac{d}{d\fwdparam}
            \log Q(\tau; \fwdparam)
        \right]
        }_{=0}
        \bigr)
    \\
    &=
    \frac{1}{S}
    \sum_{i=s}^S
    \bigl(
        -\log w_s
        + 
        c
    \bigr)
    \frac{d}{d \fwdparam}
    \log Q(\tau_s; \fwdparam) 
    .
    \end{split}
\end{align*}

\paragraph{Monte Carlo Estimation.}
We can estimate the optimal scaling with the same $S$ i.i.d.~samples $\tau_s \sim Q(\tau_s; \bwdparam)$ used to estimate $g$. However, in order for the gradient estimator to remain unbiased, we have to employ a \emph{leave-one-out} (LOO) estimator $\hat c_s$ \citep{mnih2016variational}, which only makes use of samples $\{\hat \tau_{s'} \mid s' \neq s \}$, such that 
\begin{align*}
    \mathbb{E}\left[
        \frac{1}{S}
        \sum_{i=s}^S
        \bigl(
            -\log w_s
            + 
            \hat{c}_s
        \bigr)
        \frac{d}{d \fwdparam} 
        \log Q(\tau_s; \fwdparam) 
    \right]
    =&
    \mathbb{E}\left[
        -\log w_s
        \frac{d}{d \fwdparam} 
        \log Q(\tau_s; \fwdparam) 
    \right]
    .
\end{align*}
The leave-on-out estimate of the optimal scaling for the $d$-th dimension of $c^*$ is
\begin{align*}
    \hat c^*_{d, s}
    = 
    \frac{
        \widehat{\mathrm{Cov}}_s[g_d, h_d]
    }{
        \widehat{\mathrm{Var}}_s[h_d]
    }
    ,
\end{align*}
where $\widehat{\mathrm{Cov}}_{s}[\cdot, \cdot]$, and $\widehat{\mathrm{Var}}_{s}[\cdot]$ are empirical LOO covariance and variance estimates, respectively.
Note that estimating the optimal scaling requires access to per-sample gradients and hence requires $S$ forward-backward passes on the computations graph in many reverse-mode automatic differentiation frameworks.
Two popular non-optimal scaling choices that are easily computed and do not require access to gradient information are $c^{\log w} = \mathbb{E}[\log w]$ and $c^{\log Z} = \log \mathbb{E}[w]$ with corresponding LOO estimators 
\begin{align*}
    \hat{c}^{\log w}_s
    :=  
    \frac{1}{S-1}
    \sum_{s'=1, s' \neq s}^S
    \log w_{s'}
    &&
    \hat{c}^{\log Z}_s
    := 
    \log \frac{1}{S-1}
    \sum_{s'=1, s' \neq s}^S
    w_{s'}
    .
\end{align*}
Interestingly, for $\hat{c}^{\log w}$ one can show that it is sufficient to only compute the fixed scaling $\hat{c}^{\log w}$ and instead correct by a factor $\frac{S-1}{S}$ to obtain an unbiased estimate of $g^\prime$
,
\begin{align*}
    \frac{1}{S}
    \sum_{i=1}^S
    \bigl(
        -\log w_s
        + 
        \hat{c}^{\log w}_s
    \bigr)
    \frac{d}{d \fwdparam} 
    \log Q(\tau_s; \fwdparam) 
    &=
    \frac{1}{S-1}
    \sum_{i=1}^S
    \bigl(
        -\log w_s
        + 
        \underbrace{
            \frac{1}{S}
            \sum_{j=1}^S
            \log w_j
        }_{\hat{c}^{\log w}}
    \bigr)
    \frac{d}{d \fwdparam} 
    \log Q(\tau_s; \fwdparam) 
    .
\end{align*}
In Section \ref{sec:method} we show how we can leverage these variance reduction techniques for training GFNs by identifying scenarios in which training GFNs with the TB objective is equivalent to performing variational inference with a score-function gradient estimator.

\section{Variational Inference for Generative Flow Networks}
\label{sec:method}
The trajectory balance objective and variational objectives, introduced in \ref{subsec:background_variance_reduction}, all try to find a forward model $Q$ and backward model such that
\begin{align*}
    P(\tau; \bwdparam) 
    = \pi_T(s_T) \prod_{t=0}^{T-1} P_B(s_t \mid s_{t+1}; \bwdparam)
    \approx 
    \prod_{t=0}^{T} P_F(s_{t+1} \mid s_{t}; \fwdparam) = Q(\tau; \fwdparam)
    ,
\end{align*}
and hence terminating states $s_T$ which are approximately distributed according to $\pi_T$, which is proportional to the reward $R$. 
While the TB objective can be optimized with samples from any proposal distribution that has full support on $\mathcal{T}$, it is commonly optimized with samples from either the forward model $\tau _F \sim Q$ or the backward model $\tau_B \sim P$. Similarly, variational inference commonly optimizes the RKL divergence or FKL divergence, which can be estimated by sampling from the forward model and reverse model, respectively.

\citet{zhang_generative_2022} propose a special case of the trajectory balance objective using a proposal that first samples a Bernoulli random variable $u \sim \mathcal{B}(\alpha)$. This variable then determines whether the trajectory samples are drawn from the forward model or the backward model. The corresponding expected gradient is 
\begin{align}
    \label{eq:expected_gradient_aTB}
    &\mathbb{E}_{u\sim \mathcal{B}(\alpha)}
    \left[
        [u = 0]
        \mathbb{E}_{{\tau} \sim P(\cdot; \bwdparam)}
        \left[
            \frac{d}{d\allparam}
            \mathcal{L}_{\mathrm{TB}}(\tau, \allparam)
        \right]
        +
        [u = 1]
        \mathbb{E}_{\tau \sim Q(\cdot; \fwdparam)}
        \left[
            \frac{d}{d\allparam}
            \mathcal{L}_{\mathrm{TB}}(\tau, \allparam)
        \right]
    \right]
    \\
    =&
    \alpha
    \mathbb{E}_{{\tau_B} \sim P(\cdot; \bwdparam)}
    \left[
        \frac{d}{d\allparam} 
        \mathcal{L}_{\mathrm{TB}}(\tau, \allparam)
    \right]
    +
    (1-\alpha)
    \mathbb{E}_{\tau \sim Q(\cdot; \fwdparam)}
    \left[
        \frac{d}{d\allparam}
        \mathcal{L}_{\mathrm{TB}}(\tau, \allparam)
    \right]
    .
\end{align}
We can approximate the expected gradient by approximating the expectation w.r.t.~the forward and backward model for any backward ratio $\alpha \in [0, 1]$, which is equivalent to optimizing a weighted sum of TB objectives,
\begin{align*}
    \mathcal{L}_{\alpha \mathrm{TB}}(\tau_F, \tau_B, \allparam)
    :=
    \alpha
    \mathcal{L}_\mathrm{TB}(\tau_B, \allparam)
    +
    (1-\alpha)
    \mathcal{L}_\mathrm{TB}(\tau_F, \allparam)
    ,
\end{align*}
where $\tau_F \sim Q(\cdot; \fwdparam)$ and $\tau_B \sim P(\cdot; \bwdparam)$.
We can similarly define a convex combination of the two KL divergences, which penalizes the RKL objective and FKL objective with $(1-\alpha)$ and $\alpha$, respectively, 
\begin{align*}
    \mathcal{L}_{\alpha \mathrm{KL}}(\fwdparam, \bwdparam, \alpha)
    =&
    (1 - \alpha)
    \mathcal{L}_\mathrm{RKL}(\fwdparam, \bwdparam)
    + 
    \alpha 
    \mathcal{L}_\mathrm{FKL}(\fwdparam, \bwdparam).
\end{align*}
Like RKL and FKL, this is a divergence which is non-negative and zero if and only if $P=Q$.

We are now equipped to compare the various objectives for different setting of $\alpha$ and different parameterizations of the forward and backward model. Specifically, we will differentiate between two settings: (1) the setting where $P_F$ and $P_B$ (and hence $Q$ and $P$) have distinct parameters $\fwdparam$ and $\bwdparam$ respectively, and (2) the setting where $P_F$ and $P_B$ share parameters $\sharedparam=\fwdparam = \bwdparam$.
The expected gradient of $\mathcal{L}_{\alpha\mathrm{TB}}$ can be computed as the convex combination of the expected gradient of the TB objective w.r.t. samples from the forward model and the expected gradient w.r.t.~ samples from the backward model (see Equation \ref{eq:expected_gradient_aTB}). Similarly, $\mathcal{L}_{\alpha\mathrm{KL}}$ can be computed as convex combination of $\mathcal{L}_\mathrm{RKL}$ and $\mathcal{L}_{FKL}$. Thus, in the following we study the cases $\alpha=0$ and $\alpha=1$ separately and results for $0<\alpha<1$ follow accordingly. 

\subsection{Forward model and backward model with shared parameters}
\label{ssec:shared_params}
If the forward and reverse model share parameters $\eta = (\fwdparam, \bwdparam)$, e.g. when they are parameterized by the same GFN, the expected gradient of the TB objective (Equation~\ref{eq:tb_objective}) takes the form 
\begin{align*}
    \mathbb{E}_{\tau \sim q(\cdot; \eta)}
    \left[
        \frac{d}{d \allparam}
        \mathcal{L}_{\mathrm{TB}}(\tau, \allparam)
    \right]
    &=
    -2
    \mathbb{E}_{\tau \sim q(\cdot; \eta)}
    \left[
        \left(
            \log w
            +
            \log
            \frac{Z}{Z_\zparam}
        \right)
        \left(
            \frac{d}{d \zparam}
            \log Z_\zparam
            +
            \frac{d}{d \sharedparam}
            \log Q(\tau; \sharedparam)
            -
            \frac{d}{d \sharedparam}
            \log P(\tau; \sharedparam)
        \right)
    \right]
    ,
\end{align*}
where the proposal $q$ is either the forward model $Q(\tau; \fwdparam)$ ($\alpha = 0$) or backward model $P(\tau; \bwdparam)$ ($\alpha = 1$).
The corresponding gradients of the RKL and FKL divergences are
\begin{align*}
    \frac{d}{d \sharedparam}
    \mathcal{L}_{\mathrm{RKL}}(\sharedparam)
    &=
    -
    \mathbb{E}_{\tau \sim Q(\cdot; \sharedparam)}
    \left[
        \bigl(
            \log w
            + c
        \bigr)
        \frac{d}{d \sharedparam}
        \log Q(\tau; \sharedparam)
        +
        \frac{d}{d \sharedparam}
        \log P(\tau; \sharedparam)
    \right]
    ,
    \\
    \frac{d}{d \sharedparam}
    \mathcal{L}_{\mathrm{FKL}}(\sharedparam)
    &=
    \mathbb{E}_{\tau \sim P(\cdot; \sharedparam)}
    \left[
        \bigl(
            \log w
            + c
        \bigr)
        \frac{d}{d \sharedparam}
        \log P(\tau; \sharedparam)
        -
        \frac{d}{d \sharedparam}
        \log Q(\tau; \sharedparam)
    \right]
    ,
\end{align*}
where $c$ is a scaling parameter as discussed in Section \ref{subsec:background_variance_reduction}.

\subsection{Forward model and backward model with distinct parameters}
\paragraph{Sampling from the forward model ($\alpha = 0$).}
In the case where we are using samples from the forward model $\tau \sim Q(\cdot; \fwdparam)$ only, the expected TB gradients reduce to
\begin{align*}
    \mathbb{E}_{\tau \sim Q(\cdot; \fwdparam)}
    \left[
        \frac{d}{d \fwdparam}
        \mathcal{L}_{\mathrm{TB}}(\tau,\allparam)
    \right]
    &=
    -
    2
    \mathbb{E}_{\tau\sim Q(\cdot; \fwdparam)}
    \left[
        \left(
            \log w
            +
            \cancel{
                \log
                \frac{Z}{Z_\psi}
            }
        \right)
        \frac{d}{d \fwdparam} \log Q(\tau; \fwdparam)
    \right]
    ,
    \\
    \mathbb{E}_{\tau\sim Q(\cdot; \fwdparam)}
    \left[
        \frac{d}{d \bwdparam}
        \mathcal{L}_{\mathrm{TB}}(\tau,\allparam)
    \right]
    &=
    2
    \mathbb{E}_{\tau\sim Q(\cdot; \fwdparam)}
    \left[
        \left(
            \log w
            +
            \log
            \frac{Z}{Z_\psi}
        \right)
        \frac{d}{d \bwdparam} \log P(\tau; \bwdparam)
    \right]
    ,
    \\
    \mathbb{E}_{\tau\sim Q(\cdot; \fwdparam)}
    \left[
        \frac{d}{d \zparam}
        \mathcal{L}_{\mathrm{TB}}(\tau,\allparam)
    \right]
    &=
    2
    \mathbb{E}_{\tau\sim Q(\cdot; \fwdparam)}
    \left[
        \left(
            \log w
            +
            \log
            \frac{Z}{Z_\psi}
        \right)
        \frac{d}{d \zparam} 
        \log Z_\zparam
    \right]
    .
\end{align*}
Interestingly, the expected gradient w.r.t.~$\fwdparam$ does not depend on $\log Z_\zparam$ and is proportional to the gradient of the standard score-function gradient for the reverse KL-divergence
\begin{align*}
    \frac{d}{d \fwdparam}
    \mathcal{L}_{\mathrm{RKL}}(\fwdparam, \bwdparam)
    &=
    -
    \mathbb{E}_{\tau\sim Q(\cdot; \fwdparam)}
    \left[
        \bigl(
            \log w 
            + 
            c
        \bigr)
        \frac{d}{d \fwdparam} 
        \log Q(\tau; \fwdparam) 
    \right]
    =
    \frac{1}{2}
    \mathbb{E}_{\tau\sim Q(\cdot; \fwdparam)}
    \left[
        \frac{d}{d \fwdparam}
        \mathcal{L}_{\mathrm{TB}}(\tau, \allparam)
    \right]
    .
\end{align*}
Hence, solutions of the corresponding optimization problem correspond to fixed points of the (negative) expected gradient.
Moreover, the term $\log Z/Z_\zparam$ can be interpreted as a learned scaling parameter $c_\zparam$ for variance reduction similar to the control variates discussed in section \ref{subsec:background_variance_reduction}.
Optimizing the TB objective w.r.t.~parameters of the forward model is equivalent to optimizing a RKL divergence using a score-function estimator with a learned scaling parameter $c_\zparam$, updated according to the gradient described above.
This insight also suggests that the control variate described in Section \ref{subsec:background_variance_reduction} can be used as an alternative to the learned baseline to reduce the variance of the expected gradient estimates of the trajectory balance objective.

The expression of the gradient of the RKL w.r.t.~parameters of the backward model $\bwdparam$ differs from the expected gradient of the corresponding TB objective

\begin{align*}
    \frac{d}{d \bwdparam}
    \mathcal{L}_{\mathrm{RKL}}(\fwdparam, \bwdparam)
    &=
    -
    \mathbb{E}_{\tau\sim Q(\cdot; \fwdparam)}
    \left[
        \frac{d}{d \bwdparam} 
        \log P(\tau; \bwdparam) 
    \right]
    .
\end{align*} 
The integrand differs by a multiplicative factor $\log w + c_\psi$. 

Intuitively, if the likelihood of a sample is higher under the backward transition model $P$ than under the forward transition model $Q$ by more than predicted by $-c_\zparam = \log (Z_\psi / Z)$, then $\log w + c_\zparam < 0$ and the TB objective tries to increase the likelihood of the sample under $P$ and vice versa. 
In contrast, the gradient of the RKL objective tries to always maximize the likelihood of samples under the backward transition model, which achieves its global maximum for $P = Q$. Due to the fact that $\sum_\tau P(\tau; \bwdparam) = 1$, increasing the probability of $P(\tau; \bwdparam)$ for some $\tau$ decreases the probability of other trajectories \emph{indirectly}.
Hence, while both objectives have the same global minima for flexible enough $Q$ and $P$, their optimization dynamics may differ.

\paragraph{Sampling from the backward model ($\alpha = 1$).}
When samples are taken from the backward model $\tau\sim P(\cdot; \bwdparam)$ the expected TB gradients reduce to
\begin{align*}
    \mathbb{E}_{\tau\sim P(\cdot; \bwdparam)}
    \left[
        \frac{d}{d \fwdparam}
        \mathcal{L}_{\mathrm{TB}}(\tau,\allparam)
    \right]
    &=
    -
    2
    \mathbb{E}_{\tau\sim P(\cdot; \bwdparam)}
    \left[
        \left(
            \log w
            +
            \log
            \frac{Z}{Z_\psi}
        \right)
        \frac{d}{d \fwdparam} \log Q(\tau; \fwdparam)
    \right]
    ,
    \\
    \mathbb{E}_{\tau\sim P(\cdot; \bwdparam)}
    \left[
        \frac{d}{d \bwdparam}
        \mathcal{L}_{\mathrm{TB}}(\tau,\allparam)
    \right]
    &=
    2
    \mathbb{E}_{\tau\sim P(\cdot; \bwdparam)}
    \left[
        \left(
            \log w
            +
            \cancel{
                \log
                \frac{Z}{Z_\psi}
            }
        \right)
        \frac{d}{d \bwdparam} \log P(\tau; \bwdparam)
    \right]
    ,
    \\
    \mathbb{E}_{\tau\sim P(\cdot; \bwdparam)}
    \left[
        \frac{d}{d \zparam}
        \mathcal{L}_{\mathrm{TB}}(\tau,\allparam)
    \right]
    &=
    2
    \mathbb{E}_{\tau\sim P(\cdot; \bwdparam)}
    \left[
        \left(
            \log w
            +
            \log
            \frac{Z}{Z_\psi}
        \right)
        \frac{d}{d \zparam} 
        \log Z_\zparam
    \right]
    .
\end{align*}
Here, a similar observation holds. The expected gradient, w.r.t.~$\bwdparam$, of the TB objective is proportional to the corresponding gradient of the forward KL-divergence w.r.t.~parameters $\bwdparam$
\begin{align*}
    \frac{d}{d \bwdparam}
    \mathcal{L}_{\mathrm{FKL}}(\fwdparam, \bwdparam)
    &=
    \mathbb{E}_{\tau\sim P(\cdot; \bwdparam)}
    \left[
        \log w
        \frac{d}{d \bwdparam} 
        \log P(\tau; \bwdparam) 
    \right]
    =
    \frac{1}{2}
    \mathbb{E}_{\tau\sim P(\cdot; \bwdparam)}
    \left[
        \frac{d}{d \bwdparam}
        \mathcal{L}_{\mathrm{TB}}(\fwdparam, \bwdparam, \tau)
    \right]
    .
\end{align*}
Again, solutions of the corresponding optimization problem correspond to fixed points of the (negative) expected gradient. Moreover, analogously to the previous case, optimizing the TB objective w.r.t.~$\bwdparam$ is equivalent to optimizing a FKL divergence w.r.t.~$\bwdparam$ using a score-function estimator with a learned scaling parameter $c_\zparam$. 

The expression of the gradient of the FKL w.r.t. parameters of the forward model $\fwdparam$ analogously differs from the expected gradient of the corresponding TB objective by a factor $\log w + c_\zparam$ in the integrand,
\begin{align*}
    \frac{d}{d \fwdparam}
    \mathcal{L}_{\mathrm{FKL}}(\fwdparam, \bwdparam)
    &=
    -
    \mathbb{E}_{\tau\sim P(\cdot; \bwdparam)}
    \left[
        \frac{d}{d \fwdparam} 
        \log Q(\tau; \fwdparam) 
    \right]
    .
\end{align*}

Observing the expected gradients of the TB objective and corresponding gradients of the RKL and FKL shows that in certain cases optimizing the TB objective is equivalent to variational inference using reverse or forward KL divergences. This observation also suggests that we can leverage the various variance reduction techniques for score-function estimators developed in the variational inference literature.

\section{Experiments}
We have shown that for certain settings, optimizing the $\alpha$TB objective is equivalent to optimizing the $\alpha$KL objective, in the sense that the fixed points are the same and the expected gradient of the $\alpha$TB objective is proportional to the gradient of the $\alpha$KL objective. 
In these settings we can use the variance reduction techniques for score-function gradient estimators to reduce the variance of the expected gradients of the TB objective.
In settings where optimizing the $\alpha$TB objective and $\alpha$KL objective is not equivalent, it is not immediately clear if optimizing the $\alpha$KL objective is advantageous over optimizing the $\alpha$TB objective, or vice versa. 
In the following we compare the performance of the $\alpha$TB and $\alpha$KL objective with a learned baseline $\hat c^{\fwdparam} := \log Z_\fwdparam$ or LOO baseline $\hat c^{\log Z}_s$ for different values of $\alpha$.

\paragraph{Evaluation metrics.}
If samples from the target distribution $\pi_T$ are available we can sample trajectories from the backward model conditioned on $x$. Let
\begin{align*}
    &P_B(s_{0:T-1} \mid s_T; \bwdparam)
    :=
    \prod^{T-1}_{t=0} P_B(s_t \mid s_{t+1}; \bwdparam)
    \quad \text{and} \quad
    P_F(s_{1:T-1}|s_0; \fwdparam) := \prod_{t=0}^{T-2} P_F(s_{t+1}|s_t; \fwdparam)
    .
\end{align*}
Then, we can estimate the marginal likelihood of the data under the the forward model using importance sampling,
\begin{align}
    \label{eq:ll_computation}
    \frac{1}{N} \sum_{i=1}^N 
    \frac{
            P_F(x|s_{T-1}^i; \fwdparam) P_F(s_{1:T-1}^i|s_0^i; \fwdparam)
        }{
            P_B(s_{0:T-1}^i \mid x; \bwdparam)
        }
    ,&& 
    s_{0:T-1}^i \sim P_B(s_{0:T-1} \mid x; \bwdparam)
    .
\end{align}
If no data is available we will report the expected log-weight $\mathbb{E}_{\tau \sim Q(\cdot; \fwdparam)}[\log w] \leq \log Z$.

\paragraph{Structure and representation of the state space}
Following \citet{zhang_generative_2022} we target a discrete distribution over terminating states on $\mathcal{S}_T=\{0, 1\}^{D}$ by consecutively sampling values in $\{0, 1\}$ for each step. 
To this end we define the state space $\mathcal{S} = \{\emptyset, 0, 1\}^{D} \cup \{s_f\}$, where $\emptyset$ indicates that no bit value has been sampled for the corresponding position yet. We further define edges
\begin{align*}
    E
    = 
    \{
        (s, s') 
        : 
        s \in \mathcal{S}\setminus\{s_f\} 
        \land 
        s' \in \mathcal{S'}(s)
    \}
    \cup
    \{
        (s, s_f) : s \in \mathcal{S}_T
    \}
    ,&&
    \mathcal{S}'(s) 
    =
    \{
        s' \in \mathcal{S}\setminus\{s_f\}
        : 
        \left|s\right|
        = 
        \left|s'\right| - 1
    \}
    ,
\end{align*}
where $\left|s\right|$ denotes the number of set bits in $s$. With these definitions on place we define a DAG $G(\mathcal{S}, E)$ that specifies the structure of the state space.
For mathematical convenience, we map the states $s$ to numeric representations $\tilde s$ 
in which $\emptyset$, $0$ and $1$ are replaced by $0$, $-1$ and $1$ respectively.
This allows us to compute the number of set bits $\left|s\right| = \sum_d \left|\tilde s_d\right|$, and the location and type of the bit added by a transition $s \to s'$ as the signed one-hot vector $\tilde s' - \tilde s$. We can also compute state $\neg \tilde s'(s, s') = \tilde s - (\tilde s' - \tilde s)$ that results from flipping the newly added bit in $s'$. These operations are useful for defining the transition model.

\paragraph{Transition model.}
We consider a fixed backward transition model $P_B(s_t \mid s_{t+1})$ which uniformly at random select a set bit and replaces it with $\emptyset$. The forward transition model $P_F(s_{t+1}\mid s_t; \fwdparam)$ uniformly at random selects $\emptyset$-bit and and replaces it with a bit value sampled from a Bernoulli distribution whose (logit) parameters are the output of a function $f_\fwdparam: \mathcal{S}\times\mathcal{S} \to \mathbb{R}_+$. The corresponding probability mass functions of the forward- and backward transition model are
\begin{align*}
    P_B(s_t \mid s_{t+1}) 
    = 
    \frac{1}{\left|s_{t+1}\right|}
    ,&&   
    P_F(s_{t+1} | s_t; \fwdparam) 
    = 
    \frac{1}{D - \left|s_t\right|}
    \frac{
        f_\fwdparam(\tilde s_t') 
    }{
        f_\fwdparam(\tilde s_t')
        +
        f_\fwdparam(\neg \tilde s'(s_t, s_{t+1}))
    }
    .
\end{align*}
In practice $f_\fwdparam: \mathbb{R}^{D} \to \mathbb{R}^{D\times 2}$ is a vector valued function parameterized by an Multilayer Perceptron (MLP) with weights $\fwdparam$. Given a state $s_t$, it produces $D$ pairs of logits associated with positions in the state vector. The state $s_{t+1}$ is required only to compute the position $d$ of the added bit, which is used to select the corresponding logits $f_\fwdparam(s)_d \in \mathbb{R}^2$.

\subsection{Synthetic densities}
To model a discrete target distribution $\pi_T$ over terminating states we follow \citet{dai_learning_2020, zhang_generative_2022} and discretize a continuous distribution $\pi^\mathrm{cont}_\mathrm{GT}: \mathbb{R}^2 \to \mathbb{R}^+$ into $2^{16}$ equally sized grid cells along each dimension. The cells are remapped to Gray code such that neighbouring grid cells differ in exactly one bit and the resulting pair of 16-bit vectors is concatenated to obtain a single 32-bit vector. 

We are interested in two settings: (1) Learning a forward model $Q(\tau; \fwdparam)$ such that its marginal distribution $Q_T(s_T; \fwdparam)$ approximates a fixed distribution $\pi_T(s_T)$ over terminating states, and (2) learning a forward model jointly with an energy function $\xi: \{0, 1\}^{32} \to \mathbb{R}$ such that the discretized ground truth density $\pi_\mathrm{GT} \approx \pi_T(s_T; \bwdparam) \propto \exp(-\xi(s_T, \bwdparam))$.
We optimize the energy function by maximizing the negative log-likelihood via stochastic gradient descent, interleaving gradient updates to the forward model and energy function. We approximate the gradient of the log-marginal likelihood 
\begin{align*}
    -
    \frac{d}{d\bwdparam} 
    \log \pi_T(s_T; \bwdparam)
    &=
    \frac{d}{d\bwdparam} 
    \left(
        \xi(s_T; \bwdparam) + \log Z_\bwdparam
    \right)
    =
    \frac{d}{d\bwdparam} 
    \xi(s_T; \bwdparam) 
    -
    \mathbb{E}_{s_T \sim \pi_T(\cdot; \bwdparam)}\left[
        \xi(s_T; \bwdparam)
    \right]
\end{align*}
using a contrastive divergence-based approach \citep{hinton_training_2002}, which replaces the expectation w.r.t.~$\pi_T$ with an expectation w.r.t.~the marginal distribution of a $K$-step Metropolis-Hastings (MH) chain $m(x' \mid x)$ initialized at data $x$,
\begin{align*}
    \mathbb{E}_{x \sim \mathcal{U}(\mathcal{X})}\left[
        \frac{d}{d\bwdparam} 
        \xi(x; \bwdparam) 
        -
        \mathbb{E}_{x' \sim m(x' \mid x)}\left[
            \xi(x'; \bwdparam)
        \right]
    \right]
    .
\end{align*}
The MH updates uses the GFN to construct proposals \citep{zhang_generative_2022}. For $K \to \infty$ this gradient update recovers the expected gradient of the log-marginal likelihood.

We evaluate the $\alpha$TB objective and $\alpha$KL objective for different values of $\alpha$ and two different control variates, a learned (LRN) control variate $c_\psi$ and estimated control variate $c_{\log Z}$ using a leave-one-out estimator (LOO). For each $\alpha$ we consider two settings: 1) jointly learning the energy function and parameters of the GFN, and 2) using a previously learned fixed energy function and learning parameters of the GFN only.
We find that, unsurprisingly, for $\alpha=0$, in which case optimizing the $\alpha$KL objective is equivalent to optimizing the $\alpha$KL objective with a learned control variate, both objectives perform comparably (numbers within one standard deviation) in terms of negative log-likelihood (see Table~\ref{tbl:synth_densities}).
For $0 < \alpha < 1$, both objective perform similarly, with $\alpha$TB having a slight edge over $\alpha$KL in terms of negative log-likelihood.
Interestingly, for $\alpha=1$, i.e. when sampling from backward model only, the performance of $\alpha$TB drops significantly while the performance of the $\alpha$KL objective remains stable.

\begin{figure}
    \centering
    \includegraphics[width=\textwidth]{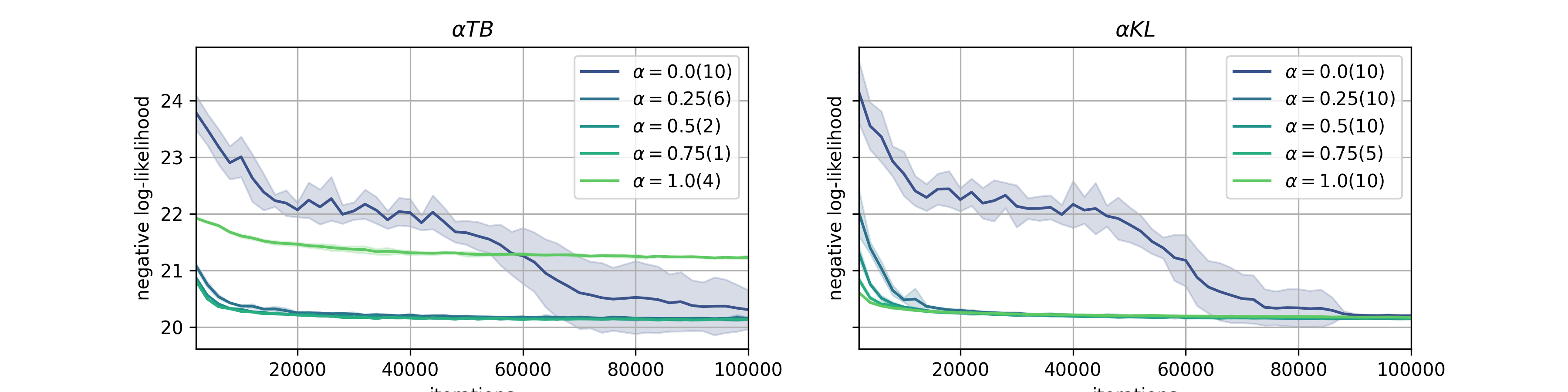}
    \caption{Negative log-likelihood during training for a fixed energy function (pre-trained on 2spirals) and different values of $\alpha$. For $\alpha=1$ the $\alpha$TB objective performs significantly worse than the $\alpha$KL divergence.}
    \label{fig:nll_over_iters}
\end{figure}

\begin{table}[b]
  \caption{Negative log-likelihood of test data under GFN for learned baseline and different backward ratio $\alpha$.}
  \label{tbl:synth_densities}
  \scriptsize
  \centering
  \begin{tabular}{lccc}
    \toprule
    Method & 2spirals & 8gaussians & 2spirals (fixed $\xi$)\\
    \midrule
    GFN $\alpha$TB (LRN, $\alpha$=0.0) & 20.163$\pm$0.013 & 20.006$\pm$0.015 & 20.307$\pm$0.343\\
    GFN $\alpha$TB (LRN, $\alpha$=0.25) & 20.133$\pm$0.010 & 20.001$\pm$0.012 & 20.156$\pm$0.017\\
    GFN $\alpha$TB (LRN, $\alpha$=0.5) & \textbf{20.118}$\pm$0.006 & \textbf{19.995}$\pm$0.008 & \textbf{20.133}$\pm$0.003\\
    GFN $\alpha$TB (LRN, $\alpha$=0.75) & \textbf{20.118}$\pm$0.009 & 20.006$\pm$0.008 & \textbf{20.133}$\pm$0.015\\
    GFN $\alpha$TB (LRN, $\alpha$=1.0) & 20.994$\pm$0.037 & 20.088$\pm$0.008 & 21.230$\pm$0.029\\
    \midrule    
    GFN $\alpha$KL (LRN, $\alpha$=0.0) & 20.171$\pm$0.015 & 20.021$\pm$0.018 & 20.200$\pm$0.015\\
    GFN $\alpha$KL (LRN, $\alpha$=0.25) & 20.142$\pm$0.012 & 19.999$\pm$0.007 & 20.153$\pm$0.009\\
    GFN $\alpha$KL (LRN, $\alpha$=0.5) & 20.145$\pm$0.008 & 20.003$\pm$0.014 & 20.147$\pm$0.012\\
    GFN $\alpha$KL (LRN, $\alpha$=0.75) & 20.160$\pm$0.008 & 20.019$\pm$0.009 & 20.155$\pm$0.011\\
    GFN $\alpha$KL (LRN, $\alpha$=1.0) & 20.174$\pm$0.009 & 20.019$\pm$0.010 & 20.172$\pm$0.008\\
    \midrule
    \bottomrule
  \end{tabular}
\end{table}

\subsection{Ising model}
\begin{figure}
    \label{fig:samples_ising}
    \centering
    \includegraphics[width=\textwidth]{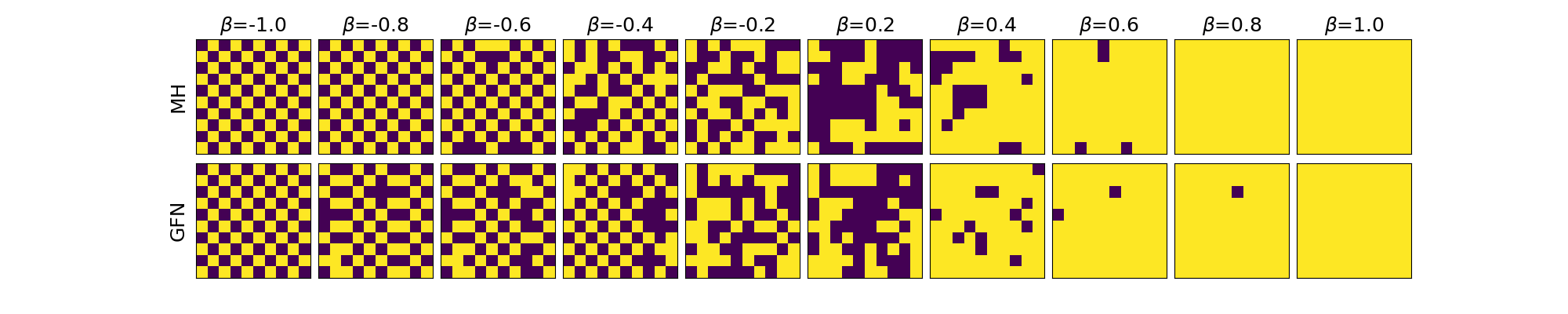}
    \caption{Approximate samples from Ising model running MH chains and forward model of a trained GFN}
\end{figure}
We are modeling a discrete distribution over terminating states $s_T \in \{-1, 1\}^D$ corresponding to the grid cells of an Ising model,
\begin{align}
    \label{eq:ising_model}
    \pi_T(s_T) \propto \exp(-\beta H(s_T)),
    &&
    H(s_T) = - \frac{1}{2} s_T^\top A_N s_T,
\end{align}
where $A$ is the adjacency matrix of a $N \times N$ ($D=N^2$) grid with periodic boundary conditions, and $\beta$ is interaction strength. In this setting, to obtain a suitable numeric representations $\tilde s$ of the states $s$ we only need to map $\emptyset$-bit to $0$. 

As we do not have access to ground truth samples from the Ising model, we are training the GFN with $\alpha=0$. In this setting, optimizing the $\alpha$TB objective and $\alpha$KL objective is equivalent and hence we focus on the effect of replacing the learned baseline, $\log Z_\fwdparam$ used in the original $\alpha$TB objective, with a $LOO$ control variate typically used to reduce the variance in score-function estimators. 

We report the expected log-weights (see Table~\ref{tbl:ising_expected_log_weight}) for different values of $\beta$ (averaged over 10 trained GFNs), and show samples from a GFN and samples generated by running a MH chain for qualitative comparison in Figure~\ref{fig:samples_ising}.
We find no significant difference in performance between the learned baseline and $LOO$ control variate across different values of $\beta$.
\begin{table}
  \caption{Expected log-weights of $\alpha$TB with different control variates for ten Ising models with different interaction strengths $\beta$. }
  \label{tbl:ising_expected_log_weight}
  \scriptsize
  \centering
  \begin{tabular}{lcccccccccc}
    \toprule
    & $\beta$=-1. & $\beta$=-0.8 & $\beta$=-0.6 & $\beta$=-0.4 & $\beta$=-0.2\\ 
    \midrule
    $\alpha$TB (LRN, $\alpha$=0.0) & \textbf{183.997}$\pm$22.010 &\textbf{ 153.512}$\pm$13.550 & \textbf{112.511}$\pm$3.967 & 42.454$\pm$1.905 & \textbf{-60.925}$\pm$0.388\\
    $\alpha$TB (LOO, $\alpha$=0.0) & 174.101$\pm$41.934 & 144.964$\pm$20.893 & 102.232$\pm$21.148 & \textbf{42.742}$\pm$1.984 & -62.970$\pm$0.276\\
    \midrule
    & $\beta$=0.2 & $\beta$=0.4 & $\beta$=0.6 & $\beta$=0.8 & $\beta$=1\\
    \midrule
    $\alpha$TB (LRN, $\alpha$=0.0) 
    & \textbf{-60.900}$\pm$0.389 & 40.707$\pm$3.733 & \textbf{112.189}$\pm$4.139 & 144.608$\pm$21.020 & 174.262$\pm$23.712\\
    $\alpha$TB (LOO, $\alpha$=0.0) 
    & -62.844$\pm$0.405 & \textbf{40.932}$\pm$1.941 & 97.109$\pm$24.809 & \textbf{153.425}$\pm$13.999 & \textbf{190.531}$\pm$19.436\\
    \midrule
    \bottomrule
  \end{tabular}
\end{table}

\section{Conclusion}

In this paper, we draw connections between the recent literature on generative flow networks and the literature on variational inference methods. We observe that GFNs can be trained using variational objectives that minimize a divergence between a forward and a backward distribution over trajectories. When minimizing the reverse Kullback-Leibler divergence, the objective is analogous to that used in standard variational inference methods that maximize a lower bound on the log-marginal likelihood \citep{blei2017variational}. When minimizing the forward Kullback-Leibler divergence, we obtain a variant of the objective that is commonly used in wake-sleep methods and related approaches \citep{hinton1995wake,bornschein2015reweighted, naesseth_markovian_2021}. It is also possible to optimize a convex combination of the two. These objectives are closely related to the trajectory-balance objective that is typically used when training GFNs. Specifically, the gradient of the RKL is proportional to computing the expected gradient of the TB objective with respect to trajectories that are sampled from the forward distribution. Evaluations on synthetic densities and an Ising model demonstrate that variational objectives for GFNs achieve a comparable performance in terms of the expected log weight relative to variants of the trajectory balance objective. This observation opens up opportunities to explore new variational objectives for GFNs that incorporate credit assignment methods \cite{schulman2015gradient} as well as importance sampling methods for GFNs based on e.g. variational sequential Monte Carlo \citep{naessethLRB2018} or nested variational inference \citep{zimmermann_nested_2021}.

\bibliographystyle{plainnat}
\bibliography{main}

\begin{thebibliography}{34}
\providecommand{\natexlab}[1]{#1}
\providecommand{\url}[1]{\texttt{#1}}
\expandafter\ifx\csname urlstyle\endcsname\relax
  \providecommand{\doi}[1]{doi: #1}\else
  \providecommand{\doi}{doi: \begingroup \urlstyle{rm}\Url}\fi

\bibitem[Bengio et~al.(2021{\natexlab{a}})Bengio, Jain, Korablyov, Precup, and
  Bengio]{bengio_flow_2021}
Emmanuel Bengio, Moksh Jain, Maksym Korablyov, Doina Precup, and Yoshua Bengio.
\newblock Flow {Network} based {Generative} {Models} for {Non}-{Iterative}
  {Diverse} {Candidate} {Generation}.
\newblock In \emph{Advances in {Neural} {Information} {Processing} {Systems}},
  volume~34, pages 27381--27394. Curran Associates, Inc., 2021{\natexlab{a}}.
\newblock URL
  \url{https://proceedings.neurips.cc/paper/2021/hash/e614f646836aaed9f89ce58e837e2310-Abstract.html}.

\bibitem[Bengio et~al.(2021{\natexlab{b}})Bengio, Deleu, Hu, Lahlou, Tiwari,
  and Bengio]{bengio_gflownet_2021}
Yoshua Bengio, Tristan Deleu, Edward~J. Hu, Salem Lahlou, Mo~Tiwari, and
  Emmanuel Bengio.
\newblock {GFlowNet} {Foundations}.
\newblock \emph{arXiv:2111.09266 [cs, stat]}, November 2021{\natexlab{b}}.
\newblock URL \url{http://arxiv.org/abs/2111.09266}.
\newblock arXiv: 2111.09266.

\bibitem[Blei et~al.(2017)Blei, Kucukelbir, and McAuliffe]{blei2017variational}
David~M Blei, Alp Kucukelbir, and Jon~D McAuliffe.
\newblock Variational inference: A review for statisticians.
\newblock \emph{Journal of the American statistical Association}, 112\penalty0
  (518):\penalty0 859--877, 2017.

\bibitem[Bornschein and Bengio(2015)]{bornschein2015reweighted}
J\"org Bornschein and Yoshua Bengio.
\newblock Reweighted wake-sleep.
\newblock In \emph{International Conference on Learning Representations}, 2015.

\bibitem[Dai et~al.(2020)Dai, Singh, Dai, Sutton, and
  Schuurmans]{dai_learning_2020}
Hanjun Dai, Rishabh Singh, Bo~Dai, Charles Sutton, and Dale Schuurmans.
\newblock Learning {Discrete} {Energy}-based {Models} via {Auxiliary}-variable
  {Local} {Exploration}.
\newblock In H.~Larochelle, M.~Ranzato, R.~Hadsell, M.~F. Balcan, and H.~Lin,
  editors, \emph{Advances in {Neural} {Information} {Processing} {Systems}},
  volume~33, pages 10443--10455. Curran Associates, Inc., 2020.
\newblock URL
  \url{https://proceedings.neurips.cc/paper/2020/file/7612936dcc85282c6fa4dd9d4ffe57f1-Paper.pdf}.

\bibitem[Del~Moral et~al.(2006)Del~Moral, Doucet, and
  Jasra]{del_moral_sequential_2006}
Pierre Del~Moral, Arnaud Doucet, and Ajay Jasra.
\newblock Sequential {Monte} {Carlo} samplers.
\newblock \emph{Journal of the Royal Statistical Society: Series B (Statistical
  Methodology)}, 68\penalty0 (3):\penalty0 411--436, 2006.
\newblock ISSN 1467-9868.
\newblock \doi{10.1111/j.1467-9868.2006.00553.x}.
\newblock \_eprint:
  https://onlinelibrary.wiley.com/doi/pdf/10.1111/j.1467-9868.2006.00553.x.

\bibitem[Deleu et~al.(2022)Deleu, Góis, Emezue, Rankawat, Lacoste-Julien,
  Bauer, and Bengio]{deleu_bayesian_2022}
Tristan Deleu, António Góis, Chris~Chinenye Emezue, Mansi Rankawat, Simon
  Lacoste-Julien, Stefan Bauer, and Yoshua Bengio.
\newblock Bayesian {Structure} {Learning} with {Generative} {Flow} {Networks}.
\newblock In \emph{The 38th Conference on Uncertainty in Artificial
  Intelligence}, June 2022.
\newblock URL \url{https://openreview.net/forum?id=HElfed8j9g9}.

\bibitem[Do et~al.(2022)Do, Dinh, Nguyen, Nguyen, Osher, and
  Ho]{do_improving_2022}
Anh Do, Duy Dinh, Tan Nguyen, Khuong Nguyen, Stanley Osher, and Nhat Ho.
\newblock Improving {Generative} {Flow} {Networks} with {Path}
  {Regularization}, September 2022.
\newblock URL \url{http://arxiv.org/abs/2209.15092}.
\newblock arXiv:2209.15092 [cs, stat].

\bibitem[Hinton(2002)]{hinton_training_2002}
Geoffrey~E. Hinton.
\newblock Training products of experts by minimizing contrastive divergence.
\newblock \emph{Neural Computation}, 14\penalty0 (8):\penalty0 1771--1800,
  August 2002.
\newblock ISSN 0899-7667.
\newblock \doi{10.1162/089976602760128018}.
\newblock URL \url{https://doi.org/10.1162/089976602760128018}.

\bibitem[Hinton et~al.(1995)Hinton, Dayan, Frey, and Neal]{hinton1995wake}
Geoffrey~E Hinton, Peter Dayan, Brendan~J Frey, and Radford~M Neal.
\newblock The" wake-sleep" algorithm for unsupervised neural networks.
\newblock \emph{Science}, 268\penalty0 (5214):\penalty0 1158--1161, 1995.

\bibitem[Hoffman(2017)]{hoffman_learning_2017}
Matthew~D. Hoffman.
\newblock Learning {Deep} {Latent} {Gaussian} {Models} with {Markov} {Chain}
  {Monte} {Carlo}.
\newblock In \emph{Proceedings of the 34th {International} {Conference} on
  {Machine} {Learning}}, pages 1510--1519. PMLR, July 2017.
\newblock URL \url{https://proceedings.mlr.press/v70/hoffman17a.html}.
\newblock ISSN: 2640-3498.

\bibitem[Hoffman and Gelman(2014)]{homan_no-u-turn_2014}
Matthew~D Hoffman and Andrew Gelman.
\newblock The {No}-{U}-{Turn} {Sampler}: {Adaptively} {Setting} {Path}
  {Lengths} in {Hamiltonian} {Monte} {Carlo}.
\newblock \emph{Journal of Machine Learning Research}, 15, 2014.

\bibitem[Jain et~al.(2022)Jain, Bengio, Garcia, Rector-Brooks, Dossou, Ekbote,
  Fu, Zhang, Kilgour, Zhang, Simine, Das, and Bengio]{jain_biological_2022}
Moksh Jain, Emmanuel Bengio, Alex-Hernandez Garcia, Jarrid Rector-Brooks,
  Bonaventure F.~P. Dossou, Chanakya Ekbote, Jie Fu, Tianyu Zhang, Micheal
  Kilgour, Dinghuai Zhang, Lena Simine, Payel Das, and Yoshua Bengio.
\newblock Biological {Sequence} {Design} with {GFlowNets}, March 2022.
\newblock URL \url{http://arxiv.org/abs/2203.04115}.
\newblock arXiv:2203.04115 [cs, q-bio].

\bibitem[Le et~al.(2018)Le, Igl, Rainforth, Jin, and Wood]{anh2018autoencoding}
Tuan~Anh Le, Maximilian Igl, Tom Rainforth, Tom Jin, and Frank Wood.
\newblock Auto-encoding sequential monte carlo.
\newblock In \emph{International Conference on Learning Representations}, 2018.

\bibitem[Li et~al.(2017)Li, Turner, and Liu]{li_approximate_2017}
Yingzhen Li, Richard~E. Turner, and Qiang Liu.
\newblock Approximate {Inference} with {Amortised} {MCMC}, May 2017.
\newblock URL \url{http://arxiv.org/abs/1702.08343}.
\newblock arXiv:1702.08343 [cs, stat].

\bibitem[Madan et~al.(2022)Madan, Rector-Brooks, Korablyov, Bengio, Jain, Nica,
  Bosc, Bengio, and Malkin]{madan_learning_2022}
Kanika Madan, Jarrid Rector-Brooks, Maksym Korablyov, Emmanuel Bengio, Moksh
  Jain, Andrei Nica, Tom Bosc, Yoshua Bengio, and Nikolay Malkin.
\newblock Learning {GFlowNets} from partial episodes for improved convergence
  and stability, September 2022.
\newblock URL \url{http://arxiv.org/abs/2209.12782}.
\newblock arXiv:2209.12782 [cs, stat].

\bibitem[Maddison et~al.(2017)Maddison, Lawson, Tucker, Heess, Norouzi, Mnih,
  Doucet, and Teh]{maddison2017filtering}
Chris~J Maddison, John Lawson, George Tucker, Nicolas Heess, Mohammad Norouzi,
  Andriy Mnih, Arnaud Doucet, and Yee Teh.
\newblock Filtering variational objectives.
\newblock In \emph{Advances in Neural Information Processing Systems}, pages
  6573--6583, 2017.

\bibitem[Malkin et~al.(2022{\natexlab{a}})Malkin, Jain, Bengio, Sun, and
  Bengio]{malkin_trajectory_2022}
Nikolay Malkin, Moksh Jain, Emmanuel Bengio, Chen Sun, and Yoshua Bengio.
\newblock Trajectory {Balance}: {Improved} {Credit} {Assignment} in
  {GFlowNets}.
\newblock \emph{arXiv:2201.13259 [cs, stat]}, January 2022{\natexlab{a}}.
\newblock URL \url{http://arxiv.org/abs/2201.13259}.
\newblock arXiv: 2201.13259.

\bibitem[Malkin et~al.(2022{\natexlab{b}})Malkin, Lahlou, Deleu, Ji, Hu,
  Everett, Zhang, and Bengio]{malkin_gflownets_2022}
Nikolay Malkin, Salem Lahlou, Tristan Deleu, Xu~Ji, Edward Hu, Katie Everett,
  Dinghuai Zhang, and Yoshua Bengio.
\newblock {GFlowNets} and variational inference, October 2022{\natexlab{b}}.
\newblock URL \url{http://arxiv.org/abs/2210.00580}.
\newblock arXiv:2210.00580 [cs, stat] version: 1.

\bibitem[Mnih and Rezende(2016)]{mnih2016variational}
Andriy Mnih and Danilo Rezende.
\newblock Variational inference for monte carlo objectives.
\newblock In \emph{International Conference on Machine Learning}, pages
  2188--2196. PMLR, 2016.

\bibitem[Naesseth et~al.(2018)Naesseth, Linderman, Ranganath, and
  Blei]{naessethLRB2018}
C.~A. Naesseth, S.~W. Linderman, R.~Ranganath, and D.~M. Blei.
\newblock Variational sequential {M}onte {C}arlo.
\newblock In \emph{Proceedings of the 21st International Conference on
  Artificial Intelligence and Statistics (AISTATS)}, Lanzarote, Spain, Apr
  2018.

\bibitem[Naesseth et~al.(2019)Naesseth, Lindsten, and Schön]{naesseth2019esmc}
C.~A. Naesseth, F.~Lindsten, and T.~B. Schön.
\newblock Elements of sequential {M}onte {C}arlo.
\newblock \emph{Foundations and Trends® in Machine Learning}, 12\penalty0
  (3):\penalty0 307--392, November 2019.
\newblock Now Publishers, Inc.

\bibitem[Naesseth et~al.(2021)Naesseth, Lindsten, and
  Blei]{naesseth_markovian_2021}
Christian~A. Naesseth, Fredrik Lindsten, and David Blei.
\newblock Markovian {Score} {Climbing}: {Variational} {Inference} with
  {KL}(p{\textbar}{\textbar}q).
\newblock \emph{arXiv:2003.10374 [cs, stat]}, February 2021.
\newblock URL \url{http://arxiv.org/abs/2003.10374}.
\newblock arXiv: 2003.10374.

\bibitem[Neal(2001)]{neal_annealed_2001}
Radford~M. Neal.
\newblock Annealed importance sampling.
\newblock \emph{Statistics and Computing}, 11\penalty0 (2):\penalty0 125--139,
  April 2001.
\newblock ISSN 1573-1375.
\newblock \doi{10.1023/A:1008923215028}.
\newblock URL \url{https://doi.org/10.1023/A:1008923215028}.

\bibitem[Ranganath et~al.(2013)Ranganath, Gerrish, and
  Blei]{ranganath_black_2013}
Rajesh Ranganath, Sean Gerrish, and David~M. Blei.
\newblock Black {Box} {Variational} {Inference}.
\newblock \emph{arXiv:1401.0118 [cs, stat]}, December 2013.
\newblock URL \url{http://arxiv.org/abs/1401.0118}.
\newblock arXiv: 1401.0118.

\bibitem[Ross(1997)]{ross_simulation_1997}
Sheldon~M Ross.
\newblock \emph{Simulation}.
\newblock academic press, 1997.

\bibitem[Salimans et~al.(2015)Salimans, Kingma, and
  Welling]{salimans_markov_2015}
Tim Salimans, Diederik Kingma, and Max Welling.
\newblock Markov {Chain} {Monte} {Carlo} and {Variational} {Inference}:
  {Bridging} the {Gap}.
\newblock In \emph{Proceedings of the 32nd {International} {Conference} on
  {Machine} {Learning}}, pages 1218--1226. PMLR, June 2015.
\newblock URL \url{https://proceedings.mlr.press/v37/salimans15.html}.
\newblock ISSN: 1938-7228.

\bibitem[Schulman et~al.(2015)Schulman, Heess, Weber, and
  Abbeel]{schulman2015gradient}
John Schulman, Nicolas Heess, Theophane Weber, and Pieter Abbeel.
\newblock Gradient estimation using stochastic computation graphs.
\newblock \emph{Advances in Neural Information Processing Systems}, 28, 2015.

\bibitem[Sutton and Barto(2018)]{sutton_reinforcement_2018}
Richard~S. Sutton and Andrew~G. Barto.
\newblock \emph{Reinforcement {Learning}, second edition: {An} {Introduction}}.
\newblock MIT Press, November 2018.
\newblock ISBN 978-0-262-35270-3.

\bibitem[Toussaint et~al.(2006)Toussaint, Harmeling, and
  Storkey]{toussaint2006probabilistic}
Marc Toussaint, Stefan Harmeling, and Amos Storkey.
\newblock Probabilistic inference for solving ({{PO}}){{MDPs}}.
\newblock \emph{Neural Computation}, 31\penalty0 (December):\penalty0 357--373,
  2006.

\bibitem[Zhang et~al.(2022{\natexlab{a}})Zhang, Chen, Malkin, and
  Bengio]{zhang_unifying_2022}
Dinghuai Zhang, Ricky T.~Q. Chen, Nikolay Malkin, and Yoshua Bengio.
\newblock Unifying {Generative} {Models} with {GFlowNets}, September
  2022{\natexlab{a}}.
\newblock URL \url{http://arxiv.org/abs/2209.02606}.
\newblock arXiv:2209.02606 [cs, stat].

\bibitem[Zhang et~al.(2022{\natexlab{b}})Zhang, Malkin, Liu, Volokhova,
  Courville, and Bengio]{zhang_generative_2022}
Dinghuai Zhang, Nikolay Malkin, Zhen Liu, Alexandra Volokhova, Aaron Courville,
  and Yoshua Bengio.
\newblock Generative {Flow} {Networks} for {Discrete} {Probabilistic}
  {Modeling}.
\newblock In \emph{Proceedings of the 39th {International} {Conference} on
  {Machine} {Learning}}, pages 26412--26428. PMLR, June 2022{\natexlab{b}}.
\newblock URL \url{https://proceedings.mlr.press/v162/zhang22v.html}.
\newblock ISSN: 2640-3498.

\bibitem[Zhang et~al.(2022{\natexlab{c}})Zhang, Blei, and
  Naesseth]{zhang2022transport}
Liyi Zhang, David~M. Blei, and Christian~A. Naesseth.
\newblock Transport score climbing: Variational inference using forward {KL}
  and adaptive neural transport.
\newblock \emph{arXiv:2202.01841}, 2022{\natexlab{c}}.

\bibitem[Zimmermann et~al.(2021)Zimmermann, Wu, Esmaeili, and van~de
  Meent]{zimmermann_nested_2021}
Heiko Zimmermann, Hao Wu, Babak Esmaeili, and Jan-Willem van~de Meent.
\newblock Nested {Variational} {Inference}.
\newblock In \emph{Advances in {Neural} {Information} {Processing} {Systems}},
  volume~34, pages 20423--20435. Curran Associates, Inc., 2021.
\newblock URL
  \url{https://proceedings.neurips.cc/paper/2021/hash/ab49b208848abe14418090d95df0d590-Abstract.html}.

\end{thebibliography}



\end{document}